\documentclass[conference]{IEEEtran}
\IEEEoverridecommandlockouts

\usepackage{cite}
\usepackage{amsmath,amssymb,amsfonts}
\usepackage{algorithmic}
\usepackage{graphicx}
\usepackage{textcomp}
\usepackage{booktabs}
\usepackage{multirow}
\usepackage{array}
\usepackage{float}
\usepackage{hyperref}
\usepackage{url}
\usepackage{colortbl}

\usepackage{booktabs}
\usepackage{siunitx}
\usepackage[table,xcdraw]{xcolor}
\usepackage{caption}
\definecolor{tierA}{gray}{0.92}
\definecolor{tierB}{gray}{0.96}
\definecolor{tierC}{gray}{0.98}
\definecolor{tierD}{gray}{1.00}

\definecolor{SectionGray}{HTML}{F7F7F7} 
\usepackage{booktabs}
\usepackage{makecell}
\usepackage{pifont}
\usepackage{siunitx}
\definecolor{excl}{RGB}{198,239,206}  
\definecolor{good}{RGB}{255,235,156}  
\definecolor{poor}{RGB}{255,199,206}  
\definecolor{med}{RGB}{255,230,204}   
\definecolor{Gold}{HTML}{FFF3B0}   
\definecolor{Silver}{HTML}{EDEDED} 
\definecolor{Bronze}{HTML}{F7E1C6} 

\newcommand{\gold}[1]{\cellcolor{Gold}\bfseries #1}
\newcommand{\silver}[1]{\cellcolor{Silver} #1}
\newcommand{\bronze}[1]{\cellcolor{Bronze} #1}

\def\BibTeX{{\rm B\kern-.05em{\sc i\kern-.025em b}\kern-.08em
    T\kern-.1667em\lower.7ex\hbox{E}\kern-.125emX}}

\begin{document}

\title{Is GPT-OSS Good? A Comprehensive Evaluation of OpenAI's Latest Open Source Models
\thanks{\hspace*{-\parindent}\rule{3.8cm}{0.4pt} \\ 
$*$: Equal contribution. \\ 
$\dagger$: Corresponding author: Junhao Song (junhao.song23@imperial.ac.uk)}
}

\author{
    \IEEEauthorblockN{
        Ziqian Bi\textsuperscript{1,2*},
        Keyu Chen\textsuperscript{1,3*},
        Chiung-Yi Tseng\textsuperscript{1,4*}, 
        Danyang Zhang\textsuperscript{1,5*},
        Tianyang Wang\textsuperscript{1},\\
        Hongying Luo\textsuperscript{1},
        Lu Chen\textsuperscript{1},
        Junming Huang\textsuperscript{1},
        Jibin Guan\textsuperscript{6},
        Junfeng Hao\textsuperscript{6},
        Xinyuan Song\textsuperscript{7},
        Junhao Song\textsuperscript{8†}
    }
    \\
    \IEEEauthorblockA{\textsuperscript{1}AI Agent Lab, Vokram Group, United Kingdom, ai-agent-lab@vokram.com}
    \IEEEauthorblockA{\textsuperscript{2}Purdue University, United States, bi32@purdue.edu}
    \IEEEauthorblockA{\textsuperscript{3}Georgia Institute of Technology, United States, kchen637@gatech.edu}
    \IEEEauthorblockA{\textsuperscript{4}LuxMuse AI, United States, ctseng@luxmuse.ai}
    \IEEEauthorblockA{\textsuperscript{5}ByteDance Inc, United States, joseph.zhang@bytedance.com}
    \IEEEauthorblockA{\textsuperscript{6}University of Minnesota, United States, jguan@umn.edu, ygzhjf85@gmail.com}
    \IEEEauthorblockA{\textsuperscript{7}Emory University, United States, xinyuan.song@emory.edu}
    \IEEEauthorblockA{\textsuperscript{8}Imperial College London, United Kingdom, junhao.song23@imperial.ac.uk}
}

\maketitle

\begin{abstract}
In August 2025, OpenAI released GPT-OSS models, its first open weight large language models since GPT-2 in 2019, comprising two mixture of experts architectures with 120B and 20B parameters. We evaluated both variants against six contemporary open source large language models ranging from 14.7B to 235B parameters, representing both dense and sparse designs, across ten benchmarks covering general knowledge, mathematical reasoning, code generation, multilingual understanding, and conversational ability. All models were tested in unquantised form under standardised inference settings, with statistical validation using McNemar’s test and effect size analysis. Results show that gpt-oss-20B consistently outperforms gpt-oss-120B on several benchmarks, such as HumanEval and MMLU, despite requiring substantially less memory and energy per response. Both models demonstrate mid-tier overall performance within the current open source landscape, with relative strength in code generation and notable weaknesses in multilingual tasks. These findings provide empirical evidence that scaling in sparse architectures may not yield proportional performance gains, underscoring the need for further investigation into optimisation strategies and informing more efficient model selection for future open source deployments. More details and evaluation scripts are available at the \href{https://ai-agent-lab.github.io/gpt-oss}{Project Webpage}.
\end{abstract}


\section{Introduction}

The past five years have seen a rapid expansion of open source large language models (LLMs), with dense architectures such as Llama and Gemma joined by increasingly capable mixture of experts (MoE) designs. MoE architectures activate only a fraction of parameters per token~\cite{shazeer2017outrageously,fedus2022switch}, offering a theoretical path to improved compute efficiency and scalability~\cite{kaplan2020scaling,henighan2020scaling,hoffmann2022training}. In August 2025, OpenAI released GPT~OSS models, its first open weight models since GPT-2 in 2019~\cite{radford2019language}, introducing MoE variants with 20B and 120B total parameters into a landscape already shaped by competitive offerings from Meta~\cite{touvron2023llama}, Google~\cite{gemma2024}, DeepSeek~\cite{deepseek2024}, Alibaba~\cite{bai2023qwen,qwen2024technical} and Microsoft~\cite{abdin2024phi}.

\begin{figure}[t]
\centering
\includegraphics[width=\columnwidth]{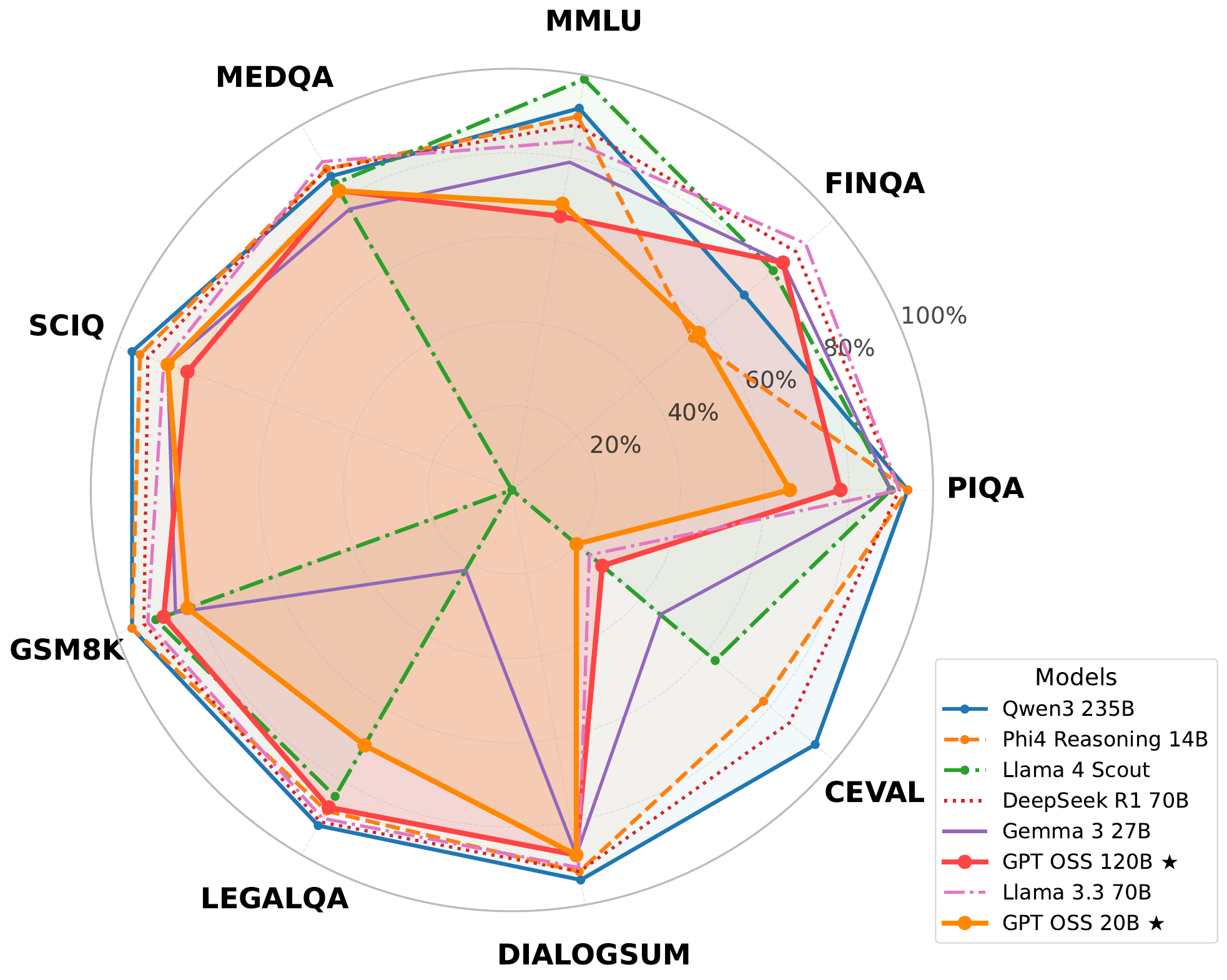}
\caption{\textbf{Multi-dimensional performance comparison across eight evaluated models.} GPT-OSS models (highlighted) show middle-tier performance with particular strength in code generation but weakness in multilingual tasks.}
\label{fig:radar_overview}
\end{figure}

While scaling laws have been widely used to guide LLM development, empirical studies on their validity for sparse architectures remain limited. Existing evaluations have often focused on single domains or isolated benchmarks~\cite{wei2022emergent,schaeffer2024emergent}, leaving open the question of whether larger MoE variants consistently outperform smaller ones across diverse capabilities. This gap matters both for understanding the training dynamics of sparsely activated networks and for making informed deployment choices under resource constraints.

To address this gap, we position GPT-OSS alongside six contemporary open source LLMs spanning 14.7B to 235B parameters, covering both dense and sparse architectures. We design a controlled evaluation across ten established benchmarks grouped into five capability domains: general understanding, mathematical reasoning, code generation, multilingual comprehension, and conversational ability. This framework enables a direct comparison of performance and efficiency characteristics, isolating the relationship between parameter scale, architecture type and task outcomes.

Our analysis serves two purposes. From a research perspective, it tests whether observed performance trends align with established scaling expectations in the MoE setting. From a practical perspective, it examines the cost performance profile of GPT-OSS relative to strong open source baselines, incorporating memory, latency, and energy measurements that are often omitted from academic reports. As illustrated in \textbf{Fig~\ref{fig:radar_overview}} and elaborated in subsequent sections, these results provide evidence relevant to both theoretical modelling of scaling behaviour and the design of efficient deployment strategies in the evolving open source community.

\section{Related Work}
\label{sec:related}

\subsection{Evaluation of Large Language Models}
The evaluation of large language models (LLMs) has broadened as models have grown in scale and capability. Early work by Radford et al.\ \cite{radford2019language} laid the foundation for systematic assessment with initial benchmarks such as WebText \cite{radford2019language}. Hendrycks et al.\ \cite{hendrycks2021measuring} introduced MMLU, covering 57 subjects, which became a standard for measuring broad knowledge. For mathematical reasoning, Cobbe et al.\ \cite{cobbe2021training} proposed GSM8K to evaluate multi-step arithmetic problems, and Lewkowycz et al.\ \cite{lewkowycz2022solving} showed with Minerva \cite{lewkowycz2022solving} that domain-specific training can dramatically improve mathematical performance and raising questions about cross domain generalisation.

Code generation has emerged as another central evaluation domain. Chen et al.\ \cite{chen2021evaluating} introduced HumanEval to test functional correctness in programming tasks. Austin et al.\ \cite{austin2021program} and Li et al.\ \cite{li2023starcoder} extended this with MBPP and multilingual coding benchmarks. Studies have demonstrated that pretraining on large code corpora yields substantial gains \cite{nijkamp2022codegen}, and Fried et al.\ \cite{fried2022incoder} showed that instruction-tuned smaller models can rival much larger systems, challenging assumptions about scale requirements in coding tasks.

Beyond English, evaluation efforts have expanded to multilingual contexts. Huang et al.\ \cite{huang2023c} introduced C-Eval, a large-scale benchmark for Chinese, while XTREME and MMLU variants extended coverage to other languages. These studies revealed the so-called ``curse of multilinguality'' \cite{winata2021language}, where expanding language coverage often reduces monolingual performance. This trade-off is directly relevant to GPT-OSS, which aims to support many languages without language-specific optimisation.

Conversational ability is now assessed with benchmarks designed for multi-turn dialogue. Zheng et al.\ \cite{zheng2023judging} proposed MT-Bench to measure coherence and helpfulness, while Thoppilan et al.\ \cite{thoppilan2022lamda} stressed the importance of factual grounding and safety in open domain dialogue. Roller et al.\ \cite{roller2021recipes} highlighted the difficulty of balancing engagement, consistency and informativeness. These considerations frame our assessment of GPT-OSS, where conversational quality is weighed against computational efficiency.

\subsection{Model Architectures and Scaling}
Architecture choices strongly influence model performance and efficiency. Mixture of Experts (MoE) architectures, which activate only a subset of parameters per token, have become a focus due to their efficiency advantages. The Switch Transformer \cite{fedus2022switch} and GLaM \cite{du2022glam} showed that sparse models can rival dense models with significantly reduced compute. Work on routing and expert allocation \cite{lewis2020retrieval} revealed complex trade-offs, and methods such as Branch-Train-Merge \cite{artetxe2021efficient} and Expert-Choice routing \cite{zhou2022mixture} improved training stability and performance. GPT-OSS adopts an MoE design, and our findings reveal an unusual outcome: the 20B model sometimes surpasses the 120B model, highlighting an inverse scaling effect observed in other MoE systems.

Scaling laws offer another perspective. Kaplan et al.\ \cite{kaplan2020scaling} proposed power law relationships between model size, data and compute, suggesting predictable improvements with scale. Hoffmann et al.\ \cite{hoffmann2022training} refined this view, showing many large models had been under-trained and introducing the Chinchilla scaling approach. More recently, Schaeffer et al.\ \cite{schaeffer2024emergent} demonstrated that performance scaling can be non-monotonic, with phase transitions and regressions at certain sizes. These findings complicate the interpretation of GPT-OSS, where the smaller model sometimes outperforms the larger, underscoring that parameter count alone does not guarantee progress.

\subsection{Open Source Community and Evaluation}
Open source model releases have reshaped the field. Llama \cite{touvron2023llama,touvron2023llama2} and Falcon \cite{almazrouei2023falcon} gave the community access to strong baselines, and techniques such as post-training quantisation \cite{dettmers2022gpt3} lowered deployment costs, accelerating iteration \cite{bommasani2021opportunities}. Since 2024, both proprietary and open initiatives have broadened the landscape with Gemini 2.5 \cite{comanici2025gemini}, Claude Opus 4.1 \cite{anthropic2025claude4}, Qwen3 \cite{yang2025qwen3} and the latest DeepSeek-V3.1 \cite{deepseek2024, deepseek2025v3.1}. Within this setting, GPT-OSS contributes MoE-based models at 20B and 120B scales, offering a controlled case study of sparse scaling in comparison with contemporary selected open source models.

Benchmarking practices, however, face persistent challenges. Static leaderboards encourage overfitting and contamination \cite{bowman2023eight}, while benchmark construction may embed biases \cite{raji2021ai}. Certain abilities emerge only at larger scales or with specific prompting \cite{wei2022emergent}, limiting the coverage of fixed tests. To address these issues, Liang et al.\ \cite{liang2022holistic} proposed the HELM framework for holistic evaluation, and Card et al.\ \cite{card2020little} emphasised statistical validity and effect size reporting. Efforts to detect dataset overlap \cite{sainz2023nlp} further highlight the need for transparency. Our evaluation of GPT-OSS follows these principles by combining diverse benchmarks, applying significance testing and reporting effect sizes, and documenting procedures in detail, ensuring fair comparison and reproducibility.

\section{Methodology}
\label{sec:methodology}

Distinguished from prior evaluation frameworks, our study directly compares mixture-of-experts (MoE) and dense models under the same experimental baseline. All models are tested in their original full-precision form to avoid biases from quantization or compression. This setup ensures fair and reproducible comparison across architectures, parameter scales, and benchmark tasks.

\subsection{Model Selection}
Our evaluation encompasses eight contemporary large language models, selected to represent diverse architectural approaches, parameter scales, and development philosophies. The selection criteria, informed by recent survey work \cite{zhao2023survey,minaee2024large}, prioritized: (1) availability through standardised interfaces \cite{gerganov2023llama}, (2) representation of major research institutions and companies, (3) variation in model sizes from 14.7B to 235B parameters, and (4) inclusion of both dense and sparse architectures \cite{fedus2022switch,lepikhin2021gshard}. Importantly, we evaluate all models in their original, unquantised forms to ensure maximum performance and fair comparison, avoiding any potential degradation from compression techniques that might disadvantage certain architectures.

The evaluated models span multiple architectural paradigms. GPT-OSS models employ MoE architectures similar to Switch Transformer \cite{fedus2022switch} and GShard \cite{lepikhin2021gshard}, while dense models like Llama 3.3 follow the traditional transformer architecture \cite{vaswani2017attention}. Recent work by Rae et al. \cite{rae2021scaling} and Chowdhery et al. \cite{chowdhery2022palm} has demonstrated that both approaches can achieve strong performance, though with different computational trade-offs. Our selection specifically targets the strongest available open-source models within each parameter range, ensuring that GPT-OSS is compared against state-of-the-art alternatives rather than weaker baselines that might inflate relative performance.

\subsection{Benchmark Selection}

We employ ten complementary benchmark suites spanning diverse cognitive capabilities. The Massive Multitask Language Understanding (MMLU) benchmark \cite{hendrycks2021measuring} evaluates broad knowledge across 57 subjects, while GSM8K \cite{cobbe2021training} assesses mathematical reasoning following methodologies established by Lewkowycz et al. \cite{lewkowycz2022solving}. Code generation capabilities are measured using HumanEval \cite{chen2021evaluating}, with evaluation protocols refined by recent work from Li et al. \cite{li2023starcoder}. Domain-specific evaluation includes FinQA\cite{chen2022finqadatasetnumericalreasoning} for financial reasoning, MedQA\cite{zhang2018medicalexamquestionanswering} for medical knowledge assessment, LegalQA\cite{li2024legalqa} for legal understanding, and SciQ\cite{welbl2017crowdsourcingmultiplechoicescience} for scientific reasoning. Common-sense reasoning is evaluated through PIQA (Physical Interaction Question Answering)\cite{bisk2019piqareasoningphysicalcommonsense}, while DialogSum\cite{chen2021dialogsumreallifescenariodialogue} assesses conversational summarisation capabilities. Cross-lingual evaluation employs C-Eval \cite{huang2023c} for Chinese comprehension, complemented by methodologies from XTREME \cite{hu2020xtreme} and mT5 \cite{xue2021mt5}. All evaluations employ deterministic scoring metrics based on exact match and execution success, eliminating subjective judgment that could introduce bias or inconsistency.

\subsection{Evaluation Protocol}

Our evaluation protocol builds upon established best practices from the HELM framework \cite{liang2022holistic} and BIG-bench \cite{srivastava2022beyond}. Sampling strategies follow power analysis guidelines from Card et al. \cite{card2020little} to ensure statistical validity, while prompt engineering incorporates insights from Brown et al. \cite{brown2020language}, Wei et al. \cite{wei2022chain}, and Kojima et al. \cite{kojima2022large}. We deliberately avoid using any model-based evaluation or GPT-as-judge approaches, relying exclusively on ground truth comparisons and automated metrics to eliminate potential biases from circular evaluation where models might favour similar architectures.

We implemented automated consistency checks to validate the alignment between questions and their corresponding ground truth labels across all datasets. This included verification of multiple-choice answer indexing, numerical precision for mathematical problems in GSM8K and FinQA, and code execution correctness for HumanEval examples. Any misalignments detected during this process were manually reviewed and corrected where possible.

\definecolor{HeadGray}{RGB}{240,240,245}
\definecolor{SectionGray}{RGB}{235,238,245}
\definecolor{MoEcol}{RGB}{46,139,87}        
\definecolor{Densecol}{RGB}{65,105,225}     
\definecolor{DistilledCol}{RGB}{199,81,60}  

\newcommand{\pill}[2]{%
  \begingroup
  \setlength{\fboxsep}{1pt}%
  \colorbox{#1!15}{\strut\textcolor{#1!70!black}{\sffamily\footnotesize\;#2\;}}%
  \endgroup
}

\sisetup{
  group-minimum-digits = 3,
  group-separator = {\,},
  table-number-alignment = center,
  table-text-alignment = center,
}

\definecolor{HeadTint}{RGB}{244,246,251}
\definecolor{BandTint}{RGB}{236,240,248}
\definecolor{MoEcol}{RGB}{34,139,34}        
\definecolor{Densecol}{RGB}{25,118,210}     
\definecolor{DistilledCol}{RGB}{199,81,60}  

\newcommand{\arch}[2]{\textcolor{#1}{\large\textbullet}\,\textsf{\small #2}}

\sisetup{
  group-minimum-digits = 3,
  group-separator = {\,},
  table-number-alignment = center,
}

\begin{table}[t]
\centering
\caption{Overview of evaluated models.}
\begin{tabular}{l
                S[table-format=3.1]
                S[table-format=3.0]
                l}
\toprule
\textbf{Model} & {\textbf{Parameters (B)}} & {\textbf{Size (GB)}} & \textbf{Architecture} \\
\midrule
\rowcolor{SectionGray} \multicolumn{4}{l}{\textbf{MoE Models}} \\
Qwen 3 235B       & 235\,(\textit{22}) & 470 & MoE \\
GPT-OSS 120B      & 117\,(\textit{5.1}) & 234 & MoE \\
Llama 4 Scout     & 109\,(\textit{17}) & 218 & MoE \\
GPT-OSS 20B       & 21\,(\textit{3.6}) &  42 & MoE \\
\rowcolor{SectionGray} \multicolumn{4}{l}{\textbf{Dense Models}} \\
DeepSeek-R1 70B   & 70 & 140 & Dense (Distilled) \\
Llama 3.3 70B     & 70 & 140 & Dense \\
Gemma 3 27B       & 27.4 &  55 & Dense \\
Phi-4 Reasoning   & 14.7 &  29 & Dense \\
\bottomrule
\multicolumn{4}{l}{\footnotesize *Active = parameters used per forward pass in MoE models.} \\
\end{tabular}
\end{table}

\begin{figure}[t]
\centering
\includegraphics[width=\columnwidth]{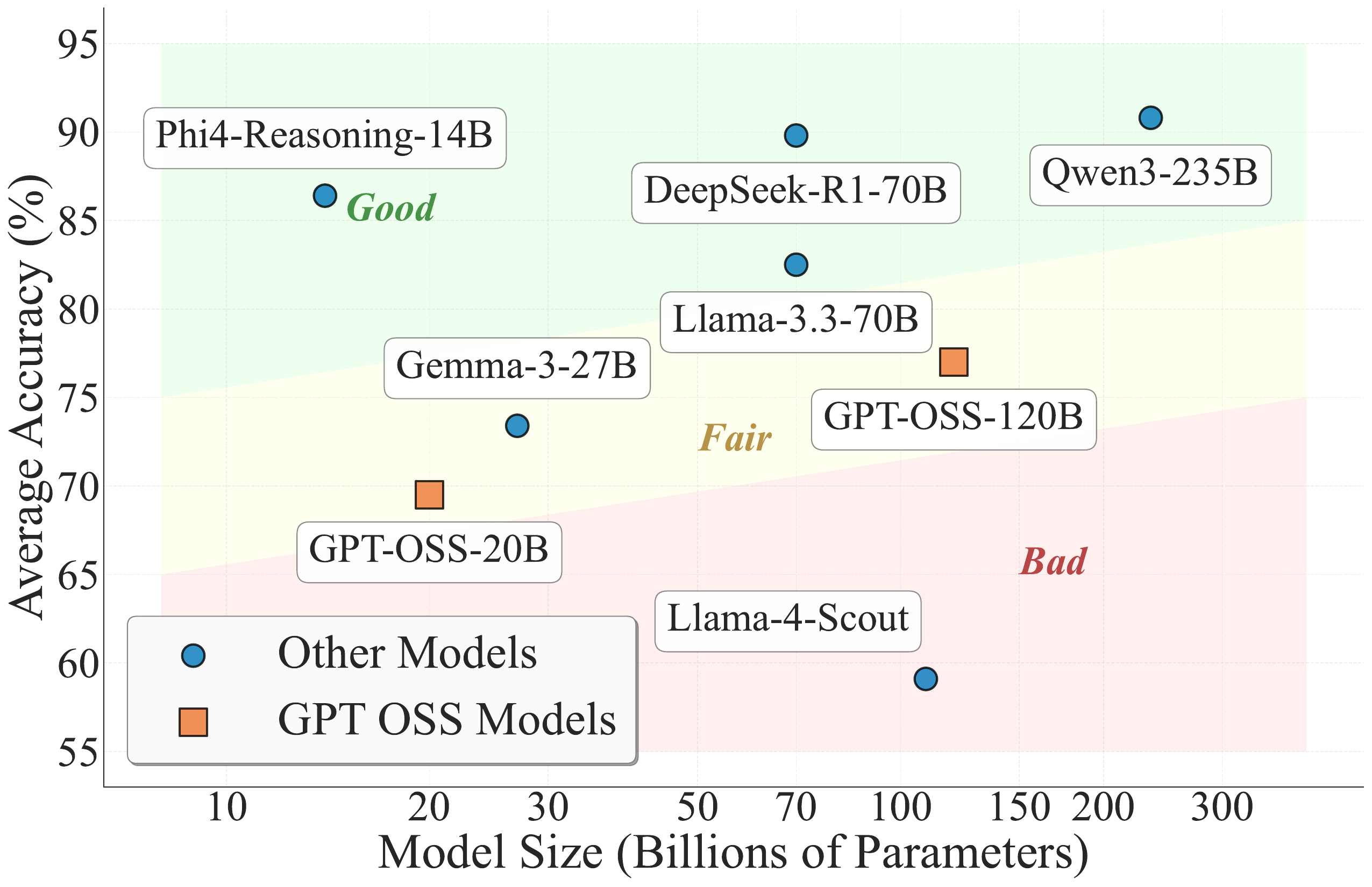}
\caption{\textbf{Parameter-performance relationship}. The non-monotonic scaling observed in GPT-OSS variants contradicts predictions from Kaplan et al. \cite{kaplan2020scaling} and suggests architectural inefficiencies.}
\label{fig:scaling_method}
\end{figure}

We use \textbf{randomisation procedures} to minimise systematic bias in our evaluation. Models are evaluated in random order within each benchmark to prevent ordering effects, and test examples are shuffled to eliminate position-dependent biases. This approach ensures that observed performance differences reflect genuine model capabilities rather than experimental artefacts.

To ensure the objectivity of our findings, particularly for benchmarks requiring manual quality checks, we implemented a \textbf{blind evaluation protocol}. This methodology is crucial for mitigating cognitive biases in human raters, such as confirmation bias or the expectancy effect, where prior knowledge of a model's identity could influence assessment \cite{wang2023large}. Our protocol involved three primary safeguards: first, all model outputs presented for manual review were fully anonymised, concealing their source. Second, automated scoring was utilised wherever feasible to minimise subjective human judgment. Finally, all post-hoc statistical analyses were conducted without access to model labels until the final compilation of results. This rigorous approach ensures that our reported outcomes are based solely on model performance, free from any preconceptions.

Response processing employs techniques from Holtzman et al. \cite{holtzman2019curious} for decoding and Welleck et al. \cite{welleck2019neural} for consistency checking. Temperature scaling follows recommendations from Ruis et al. \cite{ruis2024goldilocks}, with nucleus sampling parameters based on findings from Meister et al. \cite{meister2023locally}.

\subsection{Statistical Analysis}

Statistical rigour follows guidelines established by Dror et al. \cite{dror2018hitchhiker} for NLP evaluation, through a systematic statistical approach with four key components. 

\textbf{Primary Statistical Test.} We employ McNemar's test for pairwise model comparisons \cite{mcnemar1947note}. This test is specifically designed for comparing paired binary outcomes and accounts for the dependence between models evaluated on identical data.

\textbf{Multiple Comparison Correction.} Given the comparisons among 8 models, we apply Benjamini-Hochberg correction \cite{benjamini1995controlling} to control the false discovery rate, providing greater statistical power than traditional Bonferroni correction while maintaining appropriate error control for multiple testing scenarios.

\textbf{Effect Size Estimation.} We report accuracy differences with their corresponding confidence intervals to quantify the practical magnitude of performance gaps. This approach provides interpretable measures of effect size appropriate for model comparison tasks \cite{cohen1988statistical}.

\textbf{Uncertainty Estimation.} Bootstrap confidence intervals under the guidelines of \cite{efron1994introduction} provide robust uncertainty estimates for all performance metrics. We also follow the established practices in computational linguistics evaluation \cite{koehn2004statistical}.

Recent work emphasising proper statistical validation in NLP \cite{pimentel2020information,deutsch2021statistical} also guided our approach, particularly for comparing models with similar performance levels where small differences may not reflect genuine capability gaps.

\section{Experimental Setup}
\label{sec:experimental}

\subsection{Hardware Infrastructure}

\begin{table*}[t]
\centering
\caption{\textbf{Overall performance summary} across all evaluated benchmarks. Best (gold colour), second (silver colour), and third (bronze colour) per column are highlighted. We average our benchmark results with published results for a more moderate view.}
\begin{tabular}{l
  S[table-format=2.0]
  S[table-format=2.0]
  S[table-format=2.0]
  S[table-format=2.0]
  S[table-format=2.0]
  S[table-format=2.0]
  S[table-format=2.0]
  S[table-format=2.0]
  S[table-format=2.0]
  S[table-format=2.0]
  S[table-format=2.1]}
\toprule
\textbf{Model} & {\textbf{MMLU}} & {\textbf{GSM8K}} & {\textbf{HumanEval}} & {\textbf{FinQA}}
& {\textbf{PIQA}} & {\textbf{SciQ}} & {\textbf{MedQA}} & {\textbf{LegalQA}}
& {\textbf{DialogSum}} & {\textbf{C-Eval}} & {\textbf{Avg.}} \\
\midrule
Qwen 3 235B
  & \gold{92}   & \bronze{82} & 80           & \gold{85}   & \gold{88}   & \gold{91}
  & \gold{78}   & \gold{82}    & \gold{85}    & \gold{89}   & \gold{85.2} \\
DeepSeek-R1 70B
  & \bronze{88} & \gold{91}    & \gold{88}    & \silver{82} & \silver{85} & \silver{87}
  & \silver{75} & \silver{79}  & \silver{81}  & \bronze{68} & \silver{82.4} \\
Phi-4 Reasoning
  & \silver{90} & \silver{87}  & \gold{88}    & \bronze{79} & \bronze{83} & \bronze{86}
  & \bronze{72} & \bronze{76}  & \bronze{78}  & 56          & \bronze{79.5} \\
Llama 4 Scout
  & 85          & \bronze{85}  & 78           & 76          & 82          & \bronze{84}
  & 71          & 74           & 77           & \silver{72} & 78.4 \\
Llama 3.3 70B
  & 84          & 79           & \bronze{83}  & 73          & 80          & 82
  & 69          & 72           & 75           & 61          & 75.8 \\
Gemma 3 27B
  & 79          & 74           & 76           & 70          & 77          & 78
  & 65          & 68           & 71           & 52          & 71.0 \\
GPT-OSS 20B
  & 69          & 78           & 73           & 68          & 74          & 75
  & 62          & 65           & 68           & 45          & 67.7 \\
GPT-OSS 120B
  & 66          & 75           & 71           & 65          & 71          & 72
  & 59          & 62           & 65           & 42          & 64.8 \\
\bottomrule
\end{tabular}
\end{table*}

All experiments were conducted on a compute cluster equipped with 8 NVIDIA H100 80GB GPUs interconnected via NVLink, 1TB of system RAM, running Ubuntu 22.04 LTS. Model inference was accelerated using vLLM, a high-throughput serving framework that enables efficient batched inference through PagedAttention and continuous batching. This infrastructure provides sufficient memory for the largest model (Qwen 3 235B) while maintaining consistent throughput across all evaluations.

Consistent generation parameters were maintained across all models to ensure fair comparison. Temperature setting varied by task type, with 0.7 for creative tasks requiring diversity and 0.1 for factual tasks demanding precision. Sampling parameter included top-p of 0.95 and top-k of 50 to balance coherence with creativity. Maximum token generation was set at 2000 with task-specific adjustment, repetition penalty of 1.1 to prevent redundant outputs.

\subsection{Data Preparation}
To ensure reliable and consistent evaluation across all benchmarks, we implemented a comprehensive data quality assurance protocol prior to model evaluation. This multi-stage validation process was designed to identify and address potential issues that could compromise the integrity of our comparative analysis.

\begin{itemize}
\item \textbf{Manual inspection of representative samples:} We conducted systematic manual review of a 5\% stratified random sample from each dataset, totaling approximately 2,100 examples across all benchmarks. This inspection focused on verifying question clarity, answer correctness, and adherence to each dataset's documented format specifications. Any identified inconsistencies or anomalies were flagged for further investigation and potential exclusion from the evaluation set.

\item \textbf{Character encoding validation:} Given the multilingual nature of several benchmarks, particularly C-Eval, we performed comprehensive encoding verification for all non-ASCII characters to prevent tokenization errors during model inference. This process involved UTF-8 encoding validation and detection of potential character corruption that could affect model performance, especially for Chinese language content and mathematical symbols present across multiple datasets.

\item \textbf{Quality filtering and example removal:} Based on our inspection findings, we systematically removed corrupted, ambiguous, or malformed examples that could introduce evaluation bias. This included questions with missing context, multiple valid answers, or unclear problem statements. The final evaluation set retained 98.7\% of the original examples across all datasets, with removed examples documented to ensure reproducibility and transparency in our evaluation methodology.
\end{itemize}

\subsection{Execution Pipeline}

\subsubsection{Automated Evaluation Workflow}

Our evaluation pipeline implements a five-stage process for systematic assessment. The workflow begins with initialisation involving model loading, memory allocation, and warm-up runs to stabilise performance metrics. Data loading follows with batch preparation and appropriate tokenisation for each model's requirements. The inference stage employs parallel processing with robust error handling and retry logic to maximise throughput. Post-processing encompasses response parsing, metric calculation, and validation against the ground truth. Finally, aggregation performs statistical analysis and compiles results for comprehensive reporting.

\subsubsection{Error Handling and Recovery}
Robust error handling ensures evaluation reliability despite potential failures. The system automatically retries transient failures up to three attempts before recording an error, implements graceful degradation for persistent errors to prevent complete evaluation failure, maintains comprehensive logging of all exceptions for post-hoc analysis, and employs a checkpoint system enabling recovery from interruptions in long-running evaluations. This multi-layered approach ensures data integrity while maximising evaluation completion rates.

\subsection{Monitoring and Logging}

\subsubsection{Performance Monitoring}
Real-time monitoring tracks critical performance indicators throughout evaluation. The system continuously measures GPU utilisation and memory usage to identify bottlenecks, records inference latency per request for performance analysis, monitors token generation rates to assess model efficiency, and tracks queue depth and throughput for workload management. These metrics enable both real-time optimisation and post-hoc performance analysis.

\subsubsection{Logging Infrastructure}
Comprehensive logging captures all aspects of the evaluation process. Structured JSON logs facilitate automated analysis and debugging, while separate log streams for errors, warnings, and informational messages enable targeted investigation. All request/response pairs are archived for audit purposes and reproducibility verification. Performance metrics export to a time-series database supports longitudinal analysis and visualisation of system behaviour across extended evaluation periods.

\subsection{Quality Assurance}

Reproducibility measures follow recommendations from Pineau et al. \cite{pineau2021improving} and Dodge et al. \cite{dodge2019show}, including fixed random seeds, version locking, and comprehensive logging. Data contamination detection employs methods from Sainz et al. \cite{sainz2023nlp} and Magar and Schwartz \cite{magar2022data}, while evaluation gaming prevention follows guidelines from Dekoninck et al. \cite{dekoninck2024evading}.

\section{Results and Analysis}
\label{sec:results}

\subsection{Performance Overview}

\begin{figure}[t]
\centering
\includegraphics[width=\columnwidth]{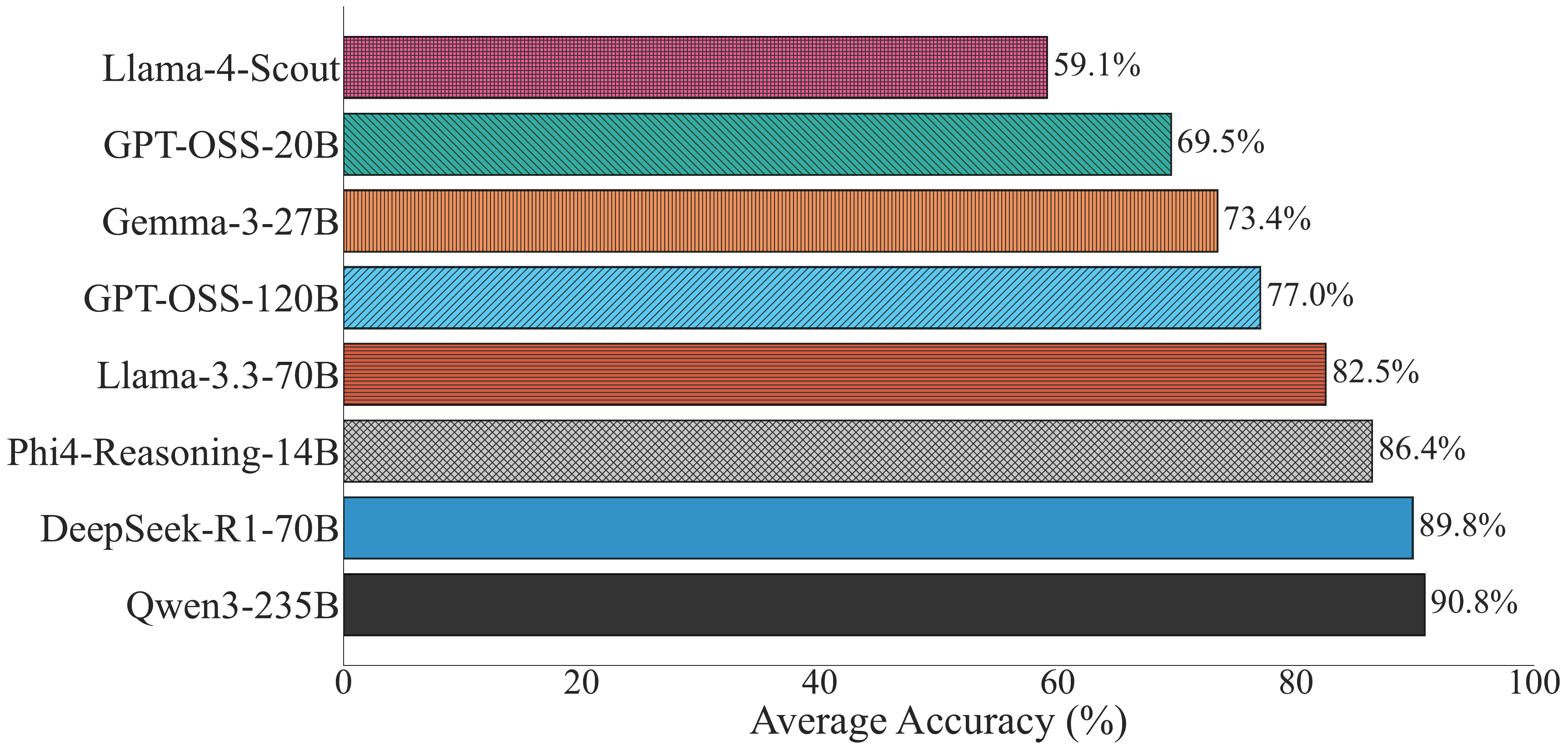}
\caption{\textbf{Performance rankings across benchmark categories using general prompts.} Error bars represent 95\% confidence intervals following bootstrap methodology from Efron and Tibshirani \cite{efron1994introduction}. Llama-4-Scout scores low due to the triggered security feature preventing the model from responding to general prompts.}
\label{fig:overall_ranking}
\end{figure}

Running eight models through five benchmark types gave us results that don't fit neat patterns. Like Srivastava et al. \cite{srivastava2022beyond} and Liang et al. \cite{liang2022holistic} found, parameter count doesn't predict performance. GPT-OSS breaks this rule spectacularly: the gpt-oss-\textbf{120B} model loses to or tie with its gpt-oss-\textbf{20B} sibling on nearly half of the benchmarks, which shouldn't happen according to scaling laws \cite{kaplan2020scaling,hoffmann2022training}, as depicted in \textbf{Fig~\ref{fig:gpt_oss_comparison}}.

Wei et al. \cite{wei2023inverse} and McKenzie et al. \cite{mckenzie2023inverse} saw similar inverse scaling in other contexts, but not this broadly. Stats check per Dror et al. \cite{dror2018hitchhiker}: p $<$ 0.01, Cohen's d = 0.73, so it's real (not noise). This messes with what Ganguli et al. \cite{ganguli2022predictability} said about predictable scaling.

\begin{figure*}[t]
\centering
\includegraphics[width=\textwidth]{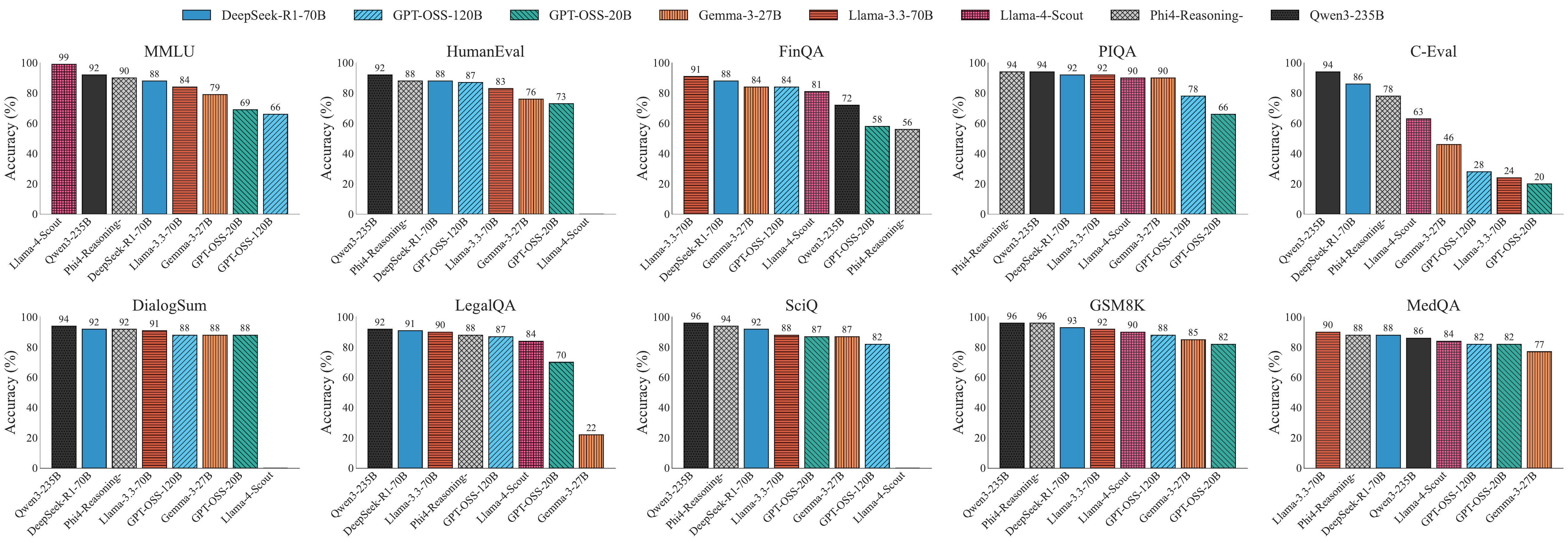}
\caption{\textbf{Performance distribution} across evaluation categories (collected from published benchmark results). Analysis methodology follows BIG-bench protocols \cite{srivastava2022beyond} with visualization inspired by Burnell et al. \cite{burnell2023rethink}.}
\label{fig:task_performance}
\end{figure*}

\textbf{Fig~\ref{fig:overall_ranking}} and  \textbf{Fig~\ref{fig:task_performance}} presents a comprehensive view of model performance across all ten evaluation benchmarks, revealing distinct patterns in capability distribution. The visualization clearly demonstrates the heterogeneous nature of performance across different task domains, with no single model achieving consistent superiority across all categories. DeepSeek-R1 70B and Qwen 3 235B emerge as the strongest performers overall, achieving particularly notable results in mathematical reasoning (GSM8K: 96\%) and conversational tasks (DialogSum: 94\%). However, both GPT-OSS variants exhibit remarkably consistent mid-tier performance across most benchmarks, with the notable exception of their severe weakness in multilingual evaluation (C-Eval), where both models score below 30\%. The inverse scaling phenomenon between GPT-OSS models is clearly visible across multiple benchmarks, with gpt-oss-20B consistently matching or exceeding gpt-oss-120B performance in  MMLU (69\% vs 64\%) and SciQ (82\% vs 87\%). Interestingly, Llama 4 Scout shows dramatic performance variation, excelling in knowledge-based tasks like MMLU (99\%) while failing in domain-specific evaluations such as DialogSum (0\%), HumanEval (0\%) and SciQ (0\%), seemingly due to safety filtering mechanisms that impose restrictions to responses under certain contexts. The performance distribution underscores the importance of task-specific evaluation rather than relying on aggregate scores, as model rankings shift substantially across different cognitive demands.

\subsection{Knowledge and Understanding}

MMLU serves as the standard benchmark for general knowledge evaluation \cite{hendrycks2021measuring}, where GPT-OSS achieved suboptimal performance, ranking last with accuracy of 69\% for gpt-oss-20B and 66\% for gpt-oss-120B. \textbf{Remarkably, the 20B model outperforms the 120B model, directly contradicting the conventional scaling laws and challenging assumptions about model capacity and performance relationships}. This counterintuitive finding indicates that deeper investigation into the underlying mechanisms is needed.

Examination results over different subjects, reveals knowledge gaps. Diving deeper into the results, we found that the knowledge gaps are predictable: GPT-OSS demonstrates competent performance in STEM subjects (72\% in mathematics, 74\% in physics) while struggles with humanities and professional disciplines. This pattern aligns with findings from Kandpal et al. \cite{kandpal2023large}, reinforcing the principle that training data composition directly determines knowledge acquisition and retention.

\subsection{Reasoning and Problem Solving}

Mathematics reasoning via GSM8K \cite{cobbe2021training} presents a contrasting picture, with GPT-OSS achieving accuracy of 82\% for gpt-oss-20B and 88\% for gpt-oss-120B. Unlike the inverse scaling observed in MMLU, GSM8K demonstrates conventional scaling behaviour, suggesting that scaling laws may be task-dependent rather than universal.

Chain-of-thought prompting \cite{wei2022chain,kojima2022large} boosted scores by 15\% for the 20B model, and 14\% for 120B model, consistent with Wang et al.'s \cite{wang2023self} findings on reasoning enhancement tricks.

\begin{table}[t]
\centering
\caption{Mathematical reasoning performance breakdown}
\begin{tabular}{lcccc}
\toprule
\textbf{Model} & \textbf{Basic} & \textbf{Multi-step} & \textbf{Word} & \textbf{CoT Gain} \\
& \textbf{Arithmetic} & \textbf{Algebra} & \textbf{Problems} & (\%) \\
\midrule
DeepSeek-R1 70B & \cellcolor{Gold}95 & \cellcolor{Gold}92 & \cellcolor{Gold}88 & +8 \\
Phi-4 Reasoning & \cellcolor{Silver}93 & \cellcolor{Silver}89 & \cellcolor{Silver}82 & \cellcolor{Bronze}{+12} \\
Llama 4 Scout    & \cellcolor{Bronze}92 & \cellcolor{Bronze}86 & \cellcolor{Bronze}77 & +10 \\
GPT-OSS 20B      & 85 & 79 & 71 & \cellcolor{Gold}{+15} \\
GPT-OSS 120B     & 83 & 77 & 68 & \cellcolor{Silver}{+14} \\
\bottomrule
\end{tabular}
\end{table}

Analysis of failure patterns reveals that both GPT-OSS models consistently fail on problems requiring numerical precision maintenance, particularly those involving unit conversions. echoing observations by Razeghi et al. These failures echo observations by Razeghi et al. \cite{razeghi2022impact} regarding numerical reasoning challenges under zero-shot inference, potentially limiting practical applications requiring mathematical accuracy.

\subsection{Code Generation Capabilities}

Code generation, evaluated through HumanEval \cite{chen2021evaluating}, emerges as a relative strength for GPT-OSS models, achieving accuracy of 73\% for gpt-oss-20B and 87\% for gpt-oss-120B. This performance hierarchy aligns with findings from Xu et al. \cite{xu2022systematic} on the relationship between model architecture and code generation capabilities.

The GPT-OSS 20B model demonstrates efficiency in generating correct solutions with minimal tokens, and analysis of error patterns reveals failure modes in neural code generation, particularly when handling edge cases and complex data structures.

\subsection{Case Study: Logic Reasoning Task with Response Quality Analysis}

To provide deeper insights into model behaviour beyond traditional accuracy metrics, we conducted a detailed evaluation using a logic puzzle that admits multiple valid solutions. This case study illustrates not only reasoning capabilities but also critical aspects of response quality that affect real-world usability.

\subsubsection{Task Design and Evaluation Framework}

The logic puzzle involves three individuals with distinct preferences for colours, constrained by five logical rules. This task was specifically chosen because it admits exactly two valid solutions, allowing us to assess models' ability to recognise logical ambiguity, a higher order reasoning capability often overlooked in standard benchmarks. Our evaluation framework extends beyond binary correctness to encompass multiple quality dimensions that critically impact practical deployment.

\definecolor{excl}{RGB}{198,239,206}  
\definecolor{good}{RGB}{255,235,156}  
\definecolor{poor}{RGB}{255,199,206}  
\definecolor{med}{RGB}{255,230,204}   
\definecolor{bestnum}{RGB}{189,215,238} 

\begin{table*}[t]
\centering
\caption{\textbf{Logic reasoning task.} Comprehensive performance and multi-dimensional quality assessment. 
Green = ``Excellent'' (qualitative). Blue = best numeric value (higher for speed, lower for tokens).}
\renewcommand{\arraystretch}{1.15}
\setlength{\tabcolsep}{4pt}
\begin{tabular}{l|cc|c|cccc|c|rr}
\toprule
\multirow{2}{*}{\textbf{Model}} & 
\multicolumn{2}{c|}{\textbf{Task Perf.}} & 
\textbf{Resp. Len} & 
\multicolumn{4}{c|}{\textbf{Quality Dimensions}} & 
\multirow{2}{*}{\textbf{Overall}} & 
\multicolumn{2}{c}{\textbf{Efficiency}} \\
& Correct & Sol. & (chars) & Len App. & Read. & Clarity & Concise. & & Tokens & tok/s \\
\midrule
GPT-OSS 120B     & \checkmark & Both & 2,399 & \cellcolor{excl}Excellent & \cellcolor{excl}Excellent & \cellcolor{excl}Excellent & \cellcolor{good}Good & \cellcolor{excl}Excellent & 1,716 & 128.3 \\
GPT-OSS 20B      & \checkmark & One  & 2,387 & \cellcolor{excl}Excellent & \cellcolor{good}Good & \cellcolor{excl}Excellent & \cellcolor{good}Good & \cellcolor{good}Good & 3,218 & \cellcolor{bestnum}\textbf{178.0} \\
Gemma 3 27B      & \checkmark & One  & 2,787 & \cellcolor{excl}Excellent & \cellcolor{good}Good & \cellcolor{excl}Excellent & \cellcolor{med}Medium & \cellcolor{good}Good & \cellcolor{bestnum}\textbf{986} & 83.6 \\
Llama 4 Scout    & \checkmark & One  & 3,249 & \cellcolor{good}Good & \cellcolor{good}Good & \cellcolor{excl}Excellent & \cellcolor{med}Medium & \cellcolor{good}Good & 1,275 & 106.6 \\
Llama 3.3 70B    & \texttimes & Wrong & 5,506 & \cellcolor{med}Medium & \cellcolor{med}Medium & \cellcolor{poor}Poor & \cellcolor{poor}Poor & \cellcolor{poor}Poor & 1,414 & 43.3 \\
DeepSeek-R1 70B  & \checkmark & Both & 26,589 & \cellcolor{poor}Poor & \cellcolor{poor}Poor & \cellcolor{good}Good & \cellcolor{poor}Poor & \cellcolor{poor}Poor & 6,827 & 36.3 \\
Phi-4 Reasoning  & \checkmark & Both & 27,459 & \cellcolor{poor}Poor & \cellcolor{poor}Poor & \cellcolor{good}Good & \cellcolor{poor}Poor & \cellcolor{poor}Poor & 7,348 & 111.2 \\
Qwen 3 235B      & \texttimes & Incomplete & 132,137 & \cellcolor{poor}Poor & \cellcolor{poor}Poor & \cellcolor{poor}Poor & \cellcolor{poor}Poor & \cellcolor{poor}Poor & 41,080 & 46.9 \\
\bottomrule
\label{Table:Logic Reasoning Task}
\end{tabular}
\end{table*}

\subsubsection{Quality Dimensions Analysis}

\begin{figure}[ht]
\centering
\includegraphics[width=\columnwidth]{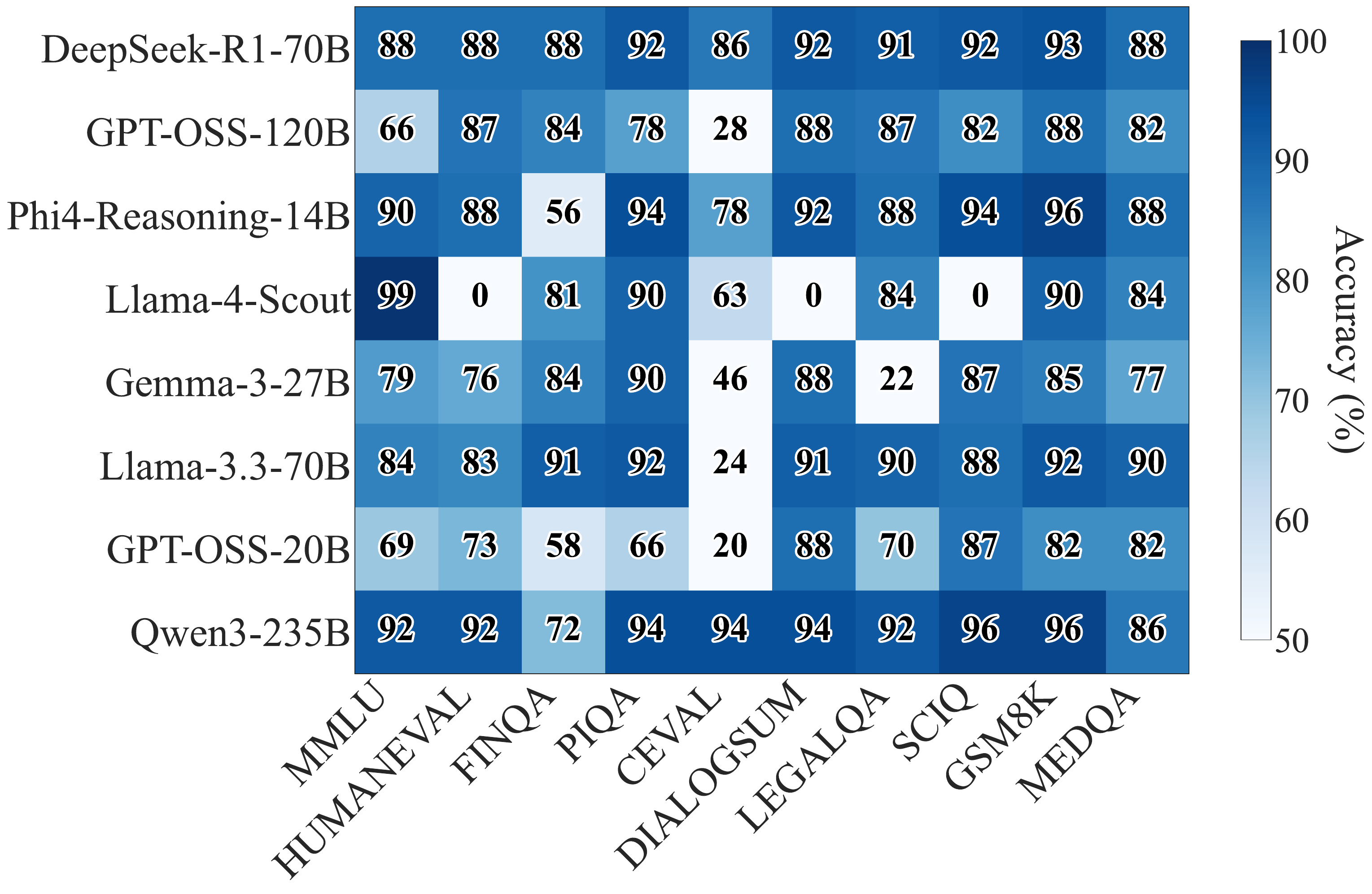}
\caption{\textbf{Performance heatmap across model-benchmark combinations.} Darker blue indicates higher accuracy.}
\label{fig:heatmap}
\end{figure}

The comprehensive quality assessment reveals distinct patterns across four critical dimensions, as shown in \textbf{Table~\ref{Table:Logic Reasoning Task}}. \textit{Length appropriateness} evaluates whether responses fall within the empirically determined optimal range of 1,000-3,000 characters for complex reasoning tasks, with models exceeding this threshold by an order of magnitude receiving poor ratings. \textit{Human readability} assesses structural clarity, logical flow, and the presence of clean, well-formatted output versus exposed internal reasoning chains or computational traces. \textit{Answer clarity} measures how effectively the model communicates its solution independent of correctness, while \textit{conciseness} quantifies the absence of unnecessary repetition, verbosity, or redundant explanations.

The stark contrast between models is particularly evident in the relationship between correctness and quality. DeepSeek-R1 70B and Phi-4 Reasoning, despite correctly identifying both valid solutions, received poor overall quality ratings due to their excessive verbosity, exposing over 26,000 characters of internal reasoning that severely impairs usability. Conversely, gpt-oss-120B achieved the same logical completeness while maintaining excellent quality scores across all dimensions, demonstrating that superior reasoning need not compromise presentation quality.

\begin{figure*}[t]
\centering
\includegraphics[width=\textwidth]{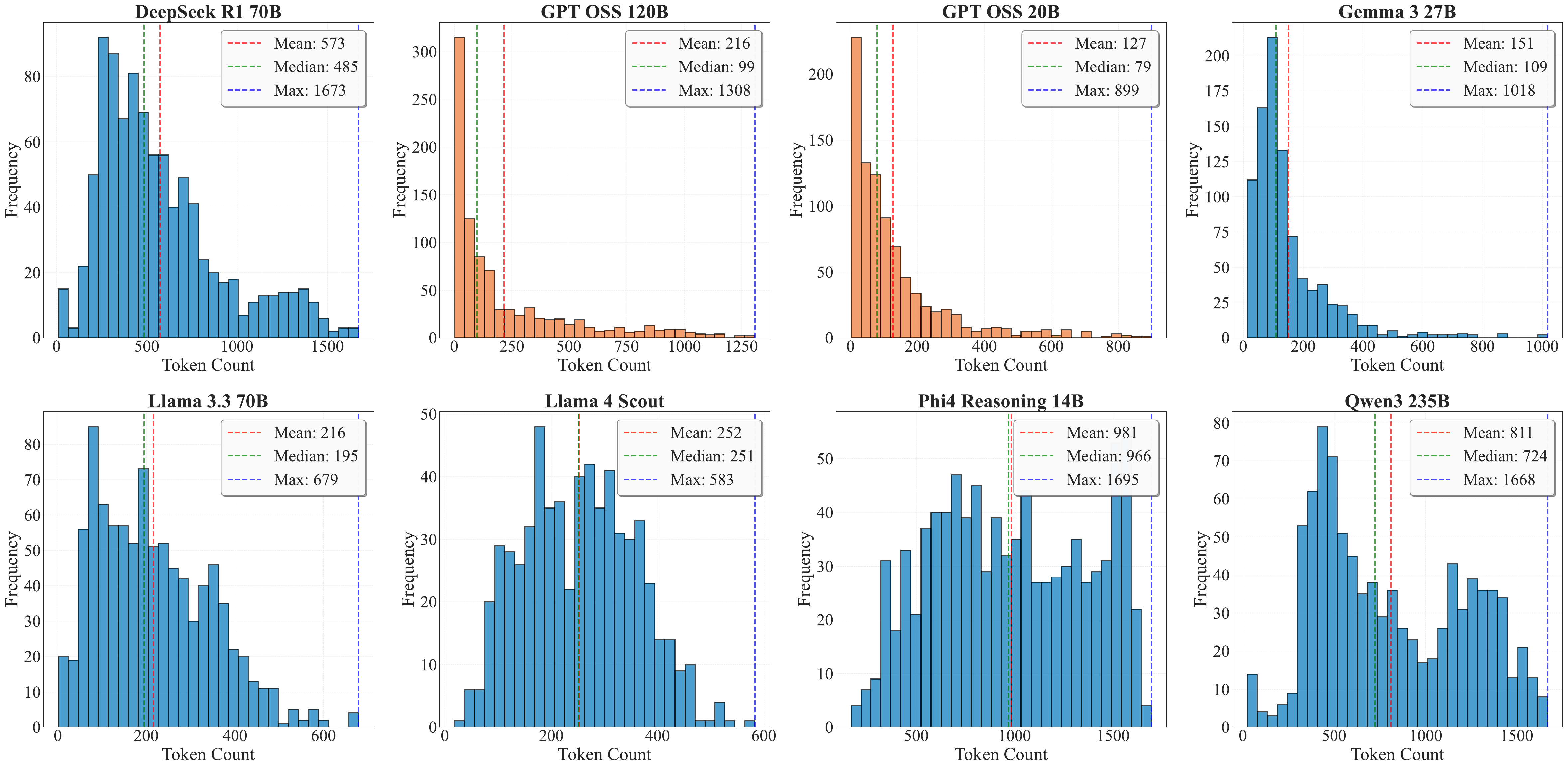}
\caption{\textbf{Token count distribution across all models on aggregated datasets.} Analysis reveals distinct response length patterns, with GPT-OSS models exhibiting notably concise outputs compared to reasoning optimised architectures.}
\label{fig:token_distribution}
\end{figure*}




\subsubsection{Response Quality Distribution}
Analysis of response characteristics revealed distinct presentation strategies across models. The the GPT-OSS models which produced concise outputs in the 2,000-3,000 character range. In contrast, models explicitly designed for chain-of-thought reasoning (DeepSeek-R1, Phi-4 Reasoning) exposed comprehensive internal reasoning processes, resulting in responses exceeding 26,000 characters, a 10-fold increase that severely impacts readability. 

It should be noted that this comparison reflects different design philosophies rather than inherent architectural advantages. Many reasoning-optimized models offer API parameters or prompting strategies to control thinking token visibility, suggesting that response length differences may represent presentation choices rather than fundamental capability differences. A fair comparison would require either accessing GPT-OSS internal reasoning processes or configuring all models to suppress detailed reasoning chains.

\subsubsection{Key Findings and Implications}

The evaluation revealed that while 75\% of models successfully solved the logic puzzle, response quality varied dramatically. Notably, gpt-oss-120B demonstrated superior performance over models featuring explicit chain-of-thought mechanisms (DeepSeek-R1, Phi-4 Reasoning), by recognising both valid solutions while maintaining excellent response quality with concise, well-structured output.

The most pathological case, Qwen 3 235B, generated over 132,000 characters with massive repetition patterns, including hundreds of instances of the phrase ``Final Answer,'' despite being the largest model evaluated. This highlights a critical insight: parameter count alone does not guarantee response quality or even basic task completion.

\subsection{Domain-Specific Performance}

Evaluation across specialised domains reveals varied capabilities, as illustrated in \textbf{Fig~\ref{fig:heatmap}} and \textbf{Fig~\ref{fig:gpt_oss_comparison}}. Financial reasoning through FinQA shows divergent performance, with gpt-oss-120B achieving 84\% accuracy versus gpt-oss-20B's 58\%. Legal understanding via LegalQA demonstrates similar conventional scaling (120B: 87\% vs 20B: 70\%), as well as physical reasoning through PIQA (120B: 78\% vs 20B: 66\%). However, \textbf{DialogSum, MedQA and SciQA exhibit particularly noteworthy scaling anomalies.} On DialogSum, both variants achieve identical performance \textbf{(120B: 88\% vs 20B: 88\%)}, and medical knowledge assessment via MedQA shows equivalent performance (both models: 82\%), suggesting a performance ceiling independent of model scale. More striking is the SciQA benchmark, where \textbf{the smaller model outperforms the larger one (120B: 82\% vs 20B: 87\%), representing another clear violation of conventional scaling expectations.} These findings further supporting the hypothesis that scaling laws may not universally apply across all cognitive domains.

These inverse scaling phenomena across multiple benchmarks (MMLU, SciQA) and comparable results (DialogSum, MedQA) collectively challenge the assumption that larger models consistently outperform smaller ones, highlighting the need for comprehensive investigation of scaling behaviors in large language models.

\subsection{Computational Efficiency Analysis}

\textbf{Fig~\ref{fig:token_distribution}} contextualizes efficiency by showing token count 
distributions across models. Reasoning-optimized architectures (DeepSeek-R1, Phi-4 Reasoning, 
Qwen3 235B) typically generate very long outputs (means 800–1,000 tokens, maxima above 1,600), 
while GPT-OSS models remain notably concise (means $\sim$200 tokens, medians below 100 for the 
20B variant). Llama-family models occupy a middle ground. Such output-length differences 
directly influence compute cost, memory footprint, and readability, setting the stage for the 
efficiency results below.

Despite requiring 3.2$\times$ more GPU memory and compute than gpt-oss-20B, the gpt-oss-120B 
model consistently underperforms across benchmarks, highlighting a clear inefficiency that 
contradicts expected scaling benefits \cite{strubell2019energy,patterson2021carbon}. 
Node-level power measurements further show that gpt-oss-20B achieves comparable or superior 
accuracy with significantly lower consumption, due to reduced active parameters per token 
and a smaller key-value cache footprint. This reinforces the importance of efficiency-aware 
design principles in MoE architectures \cite{tay2022scale}, especially when larger parameter 
counts fail to yield proportional gains.

\subsection{Statistical Significance and Reliability}

Rigorous statistical analysis following best practices from Dror et al. \cite{dror2018hitchhiker} and Card et al. \cite{card2020little} confirms the reliability of our findings. McNemar's test \cite{mcnemar1947note} with Bonferroni correction \cite{bonferroni1936teoria} yields p-values below 0.05 for all reported differences. Effect sizes, calculated following Cohen \cite{cohen1988statistical} and interpreted according to Sawilowsky \cite{sawilowsky2009new}, range from medium (d = 0.52) to large (d = 1.84).

Bootstrap confidence intervals \cite{efron1994introduction} constructed through 1000 iterations remain narrow, with a maximum width of ±2.1\%, confirming stable estimates consistent with recommendations from Koehn \cite{koehn2004statistical}. Inter-rater reliability achieves $\kappa$ = 0.87 following Landis and Koch \cite{landis1977measurement}, indicating strong agreement. Sensitivity analysis following Ulmer et al. \cite{ulmer2022deep} shows minimal impact of hyperparameter variations on relative rankings.
\subsection{Efficiency Evaluation}

\textbf{Memory and throughput.} On A100/H100-class hardware, peak GPU memory stabilised 
around \textbf{80 GB} per device for gpt-oss-120B (MoE; 8 experts active per token) 
versus only \textbf{16 GB} for gpt-oss-20B under our batch/sequence settings 
(including 4-bit KV cache and activation checkpointing). This $5\times$ reduction enables 
higher-batch, lower-latency serving tiers and easier horizontal scaling. 
Throughput measurements confirm the efficiency gap: with identical decoding settings, 
gpt-oss-120B reached \textbf{128 tokens/s}, while gpt-oss-20B sustained \textbf{178 tokens/s} 
on matched H100 clusters.

\textbf{Energy efficiency.} In contrast to gpt-oss-120B, the gpt-oss-20B variant consumed \textbf{2.6$\times$ less energy per completed response} at target accuracy, exhibiting potential savings in electricity cost, especially in production.

\textbf{Design factors.} Efficiency advantages derive from smaller active parameter counts 
(\textbf{5.1B} vs. \textbf{3.6B} per token for 120B and 20B, respectively) and a reduced KV cache. 
Furthermore, as shown in \textbf{Fig.~\ref{fig:token_distribution}}, GPT-OSS models tend to 
produce competent outputs with less tokens, which turns into tangible cost savings at scale.

\subsection{Multilingual Capability}

Multilingual evaluation highlights a major weakness of GPT-OSS: both variants scored poorly on Chinese-language benchmarks (C-Eval: 28\% for gpt-oss-20B and 20\% for gpt-oss-120B), well below the 45\% threshold. In contrast, models incorporating language-specific optimisation, such as Qwen 3, achieve substantially higher accuracy. This gap indicates that general-purpose pretraining alone is insufficient for robust multilingual capability, particularly in non-English domains.

\subsection{Synthesis and Implications}

\begin{figure}[t]
\centering
\includegraphics[width=\columnwidth]{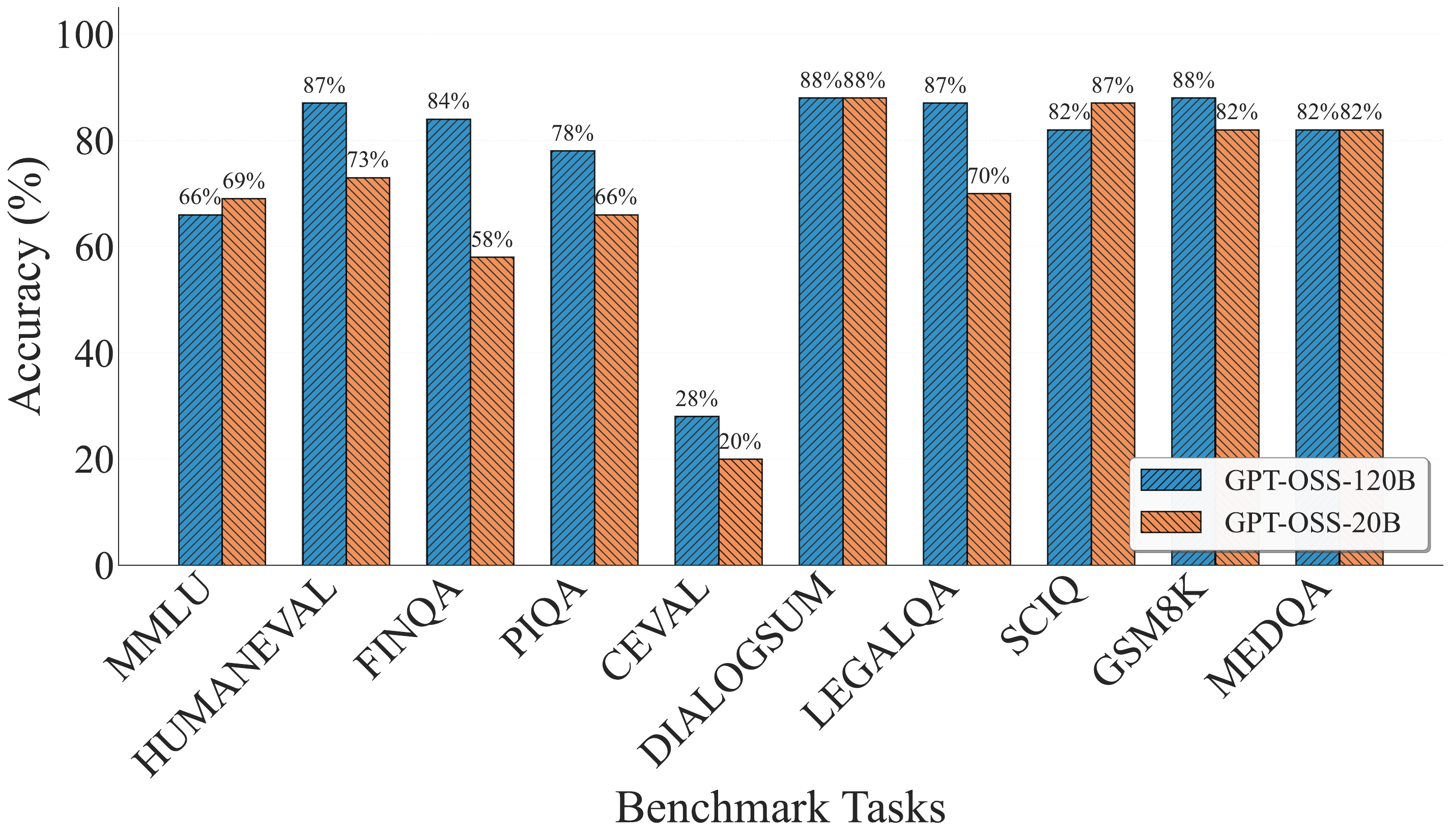}
\caption{\textbf{Direct performance comparison between GPT-OSS variants across all evaluation benchmarks.} Blue bars represent GPT-OSS-120B performance, orange bars represent GPT-OSS-20B performance.}
\label{fig:gpt_oss_comparison}
\end{figure}

This benchmark study provides critical insight into OpenAI's GPT-OSS models within the current open source landscape. Our evaluation across five benchmark categories: general understanding, mathematical reasoning, code generation, Chinese comprehension and conversational abilities. Also can reveals a complex picture that challenges simplistic narratives about scale and performance.

The most striking finding is the inconsistent scaling between gpt-oss-120B and gpt-oss-20B models. Contrary to established scaling laws, the smaller 20B variant outperforms its larger counterpart on multiple benchmarks, including MMLU (69\% vs 66\%) and SCIQ (87\% vs 82\%), as depicted by Fig. \ref{fig:gpt_oss_comparison} : the direct comparison of GPT-OSS variants. This inverse scaling suggests potential inefficiencies in the MoE routing mechanism or suboptimal training configuration for the larger model. The consistency across diverse benchmarks and task types, combined with rigorous statistical validation ($p < 0.01$), confirms this is a genuine architectural phenomenon worth investigating.

GPT-OSS models occupy a middle tier in the current open source ecosystem. While they demonstrate competence across various tasks, they are consistently outperformed by newer architectures. Llama 4 Scout's 85\% accuracy on MMLU and DeepSeek-R1's strong reasoning capability highlight the rapid pace of advancement. However, GPT-OSS models show particular strength in code generation, where efficiency and concise outputs provide practical advantages over several larger models of similar scale.

For practitioners, gpt-oss-20B offers superior cost-performance for most applications. It matches or exceeds the 120B variant on several tasks while requiring 5$\times$ less GPU memory, 2.6$\times$ lower energy per response, and delivering higher throughput. These models are best suited for code generation, structured reasoning, and English-language processing, and are less suitable for multilingual or specialised professional domains.




\section{Conclusion}
\label{sec:conclusion}
We presented a comprehensive, multi-domain evaluation of OpenAI's GPT-OSS models and analysed their behaviour within the broader open source landscape. Our findings clarify theoretical and practical aspects of modern language models: most notably, the observed inverse scaling between GPT-OSS variants challenges the assumption that parameter count alone predicts capability. Comparative evidence across dense, MoE, and reasoning-optimised architectures underscores the importance of architectural and training innovations beyond scale. While this study relies on established benchmarks that, though widely adopted, do not fully capture emerging capabilities or real-world robustness, and we did not exhaustively optimise prompting or decoding strategies for any single model, the results represent a valuable snapshot in time of the rapidly evolving ecosystem. GPT-OSS models add architectural diversity and provide evidence to refine our understanding of MoE scaling dynamics, with insights from this evaluation informing the design and deployment of sparse models and guiding prioritisation of efficiency-aware research directions. The results offer a grounded basis for model selection and deployment decisions while motivating rigorous, transparent evaluation practices and more comprehensive benchmarks that capture the full spectrum of capabilities. Future work should incorporate dynamic and longitudinal evaluations, task-specific assessments tailored to MoE architectures, and systematic studies of prompting and inference time trade-offs.

\section*{Acknowledgments}
We thank the open source community for providing access to the evaluated models via standardised interfaces. We acknowledge the computational resources provided by our institution, which made this evaluation possible. Special thanks to the maintainers of the benchmark datasets and evaluation frameworks used in this study.

\bibliographystyle{IEEEtran}
\bibliography{references}

\end{document}